# THE RATIONAL AND COMPUTATIONAL SCOPE OF PROBABILISTIC RULE-BASED EXPERT SYSTEMS

Shimon Schocken

Decision Sciences Department
The Wharton School, University of Pennsylvania

## ABSTRACT

Belief consolidation schemes in AI can be viewed as three dimensional languages, consisting of a *syntax* (e.g. probabilities or Certainty Factors), a *calculus* (e.g. Bayesian rules or CF combination rules), and a *semantics* (i.e. cognitive interpretations of competing formalisms). This paper studies the underlying rationality of those languages on the syntax and calculus grounds. In particular, the paper presents the *CF-Bayesian Endomorphism Theorem*, which explicitly defines the class of real-life problems of expertise that have a rational CF interpretation. Some implications of the theorem to the relationship between the CF and the Bayesian languages and the Dempster-Shafer Theory of Evidence are presented. Finally, the paper discusses the implications of the theorem on the science and practice of rule-based knowledge engineering.

## 1. INTRODUCTION

In order for a computer program to be a plausible model of a (more or less) rational process of human expertise, the program should be capable of representing beliefs in a language that is (more or less) calibrated with a well-specified normative criterion, e.g. the axioms of Subjective Probability [15], the Theory of Confirmation [4], formal Logic, etc. According to Shafer & Tversky, the building blocks of a probabilistic language are syntax, calculus, and semantics [18]. The **syntax** is a set of numbers, commonly referred to as Degrees of Belief (e.g. standard probabilities or Certainty Factors), which are used to parameterize uncertain facts, inexact rules, and competing hypotheses. Typically, a set of atomic Degrees of Belief (DOB's) are elicited directly from a human expert, while compound DOB's are computed through a set of operators collectively known as a belief **calculus**. Finally, the **semantics** of the language can be viewed as a mapping from a real-life domain of expertise onto the probabilistic language. This mapping provides a cognitive interpretation as well as descriptive face-validity to both the syntax and the calculus dimensions of the language.

Given the critical role that a probabilistic language plays in determining both the low-level mechanics and the high-level ordinal ranking of the recommendations of an expert system, it is clear that the implicit rationality of the language is directly related to both the internal and external validities of computer based expertise. By 'rationality' we refer here to the normative criteria of *Consistency* and *Completeness* (e.g. de Finetti, Savage [15]), as well as the psychometric criteria of *Reliability* and *Validity* [Wallsten,21]. We argue that the performance of any expert, whether a human being or a computer program, should be evaluated and rated along those lines.

The two mainstream probabilistic languages in AI are the normative *Subjective Bayesian school* and the *Certainty Factors (CF)* 'movement,' the latter being representative of a wide variety of descriptive calculi of uncertainty. It seems that the CF method is currently the most widely used probabilistic language in applied expert systems, primarily due to the popularity of such CF-based shells as EMYCIN, M.1 and Texas Instrument's Personal Consultant. Bayesian inference has been traditionally much less popular, with the exception of some notable examples, e.g. PROSPECTOR [7]. Recently, new heuristic techniques designed to cope with the computational complexity of a complete Bayesian design are emerging, giving rise to the neo-Bayesian concept of heuristic 'Bayesian inference nets' [Pearl,13].

Notwithstanding the critical importance of exploring the rational boundaries of expert systems, few studies have compared the CF and Bayesian languages on rational as well as cognitive grounds. Furthermore, there exist a number of commonly held misconceptions regarding the state of the art of AI-based belief representation techniques, such as the following two conjectures:

C1: Classical Bayesian methods are either too simplistic or too complex: in order for a Bayesian updating procedure to be computationally feasible, strict statistical independence must prevail. This requirement is rarely met in practice, where interaction effects among clues and hypotheses make the Bayesian solution unmanageable on combinatorial grounds. The CF calculus, on the other hand, does not explicitly require that the facts/hypotheses space be statistically independent, and hence can be used to model realistically complicated problems that defy a normative Bayesian interpretation.

C2: Both the Bayesian and the CF calculi are special cases of the general Dempster-Shafer Theory of Evidence [10]. Hence, they should be construed as two alternative and competing probabilistic languages, each specialized to deal with a particular class of problems and probabilistic designs.

Toward the end of the paper we will suggest a new and rather surprising interpretation of both C1 and C2. The organization of the paper is as follows: Section 2 presents an axiomatic description of the notion 'domain of expertise.' Section 3 repeats the original probabilistic interpretation of CF's, as given by Shortliffe and Buchanan in [19]. Section 4 presents three lemmas that are further integrated into the *CF-Bayesian Endomorphism Theorem* (all proofs, along with two additional lemmas, may be found in [16], which is available from the author). Sections 5 and 6 discuss the implications of the

223

theorem on the science and practice of rule-based knowledge Engineering. Section 7 concludes with future research directions.

The Endomorphism Theorem can be viewed as an integrated theory that consolidates many features of Adams [1], Heckerman [12], Prade [14], and Shortliffe & Buchanan [19]. At the same time, unlike the ad-hoc nature of some of those contributions, the normative premise of this very general and domain-independent theorem requires no more than the standard Savage/de Finetti axioms of subjective probability [15].

The contribution of the CF-Bayesian Endomorphism Theorem stems from three propositions that will be addressed in what follows:

- The theorem explicitly defines the rational boundaries of probabilistic rule-based expert system architectures.

- The theorem shows that the commonly believed conjectures C1 and C2 are misconceptions.

- The theorem points to new research directions that may promote a more rational practice of knowledge engineering.

## 2. AXIOMATIC DEFINITION OF THE 'DOMAIN OF EXPERTISE'

A 'domain of expertise' (e.g. medical diagnosis) is viewed here as a subset V of an n-dimensional propositional space S, where each dimension is interpreted as an attribute of the domain (e.g. chest pain, headache, allergy, etc.). Furthermore, each attribute can be instantiated as either an hypothesis (e.g. a disease) or as a piece of evidence (e.g. a symptom). This non-unique interpretation, which may vary across experts and contexts, is termed a 'subjective instantiation' of the space V.

We also assume that there exists a (possibly transcendental) joint distribution function F, $F : V \rightarrow [0,1]$. Although we don't have direct access to F, there exists a domain expert who is capable of making judgments that can be further interpreted as a subjective function BEL which approximates F. $BEL : 2^S \rightarrow R$ is a function of F, W, E, and L, defined as follows: W is the set of all beliefs and experiences of the human domain expert. E is the elicitation procedure designed to unravel those beliefs. L is a probabilistic language designed to encode elicited beliefs.

In order to avoid the apologetic debate of whether or not the function F exists, we note that F is presented here primarily for the sake of clear exposition. In fact, the relationship between BEL and F is at the center of an intensified philosophical debate that has been going strong for more than 300 years. In short, under a Bayesian interpretation (e.g. Ramsey), BEL = P ≡ F, where P is the standard Savage/de Finetti subjective probability function. Objectivists (like Popper) argue that F stands aloof from P (or, for that matter, from any personal BEL) and, hence, in general, P ≠ F. Logisicists (e.g. Carnap) would operationalize F through the Theory of Confirmation [4] and its contemporary AI interpretation BEL = CF.

Finally, pragmatic Bayesians (like myself) feel that BEL = P is our best shot at F, a shot whose accuracy is directly related to the operational characteristics of the elicitation procedure designed to construct P. Since P is subject to an internal axiomatic system, we term |BEL - P| an 'internal bias' and |BEL - F| an 'external bias.' Attempts to reduce those biases are termed 'debiasing procedures' or 'corrective procedures' in the cognitive psychology literature [9,11].

We now present a set of definitions that partition V into three classes of problems that vary in terms of their computational (and cognitive) complexity. In other words, we observe that some problems that require expertise may be simple 'open and shut' cases, while other problems may be complicated and vague. In what follows, We wish to provide a more precise definition of this taxonomy of problems, based on the underlying complexity of their diagnostic structures.

## DEFINITIONS

Let S be an n+1 propositional space $(X_0, X_1, \ldots, X_n)$ defined over the set (false,unknown,true)$^{n+1}$.

$X = (h, e_1, \ldots, e_n)$ is a subjective instantiation of $S = (X_0, X_1, \ldots, X_n)$ if X is a relabeling of S. Namely, $X_i = h$ for some $0 \leq i \leq n$, and $X_i = e_j$ for $j \neq i$. Without loss of generality, we assume that i=0.

A subjective instantiation $X = (h, e_1, \ldots, e_n)$ of S may be characterized by two conditionings of BEL, namely $BEL(h|e_1, \ldots, e_n)$ and $BEL(e_1, \ldots, e_n|h)$, termed the predictive solution and the diagnostic structure of X, respectively.

- X is a problem in S if

    1. X is a subjective instantiation of S.
    2. X has a diagnostic structure and a (possibly unknown) predictive solution.

- The Domain of Expertise V:

    $V = \{X \mid X \text{ is a problem in } S\}$

- The set of Weakly Decomposable Problems WD:

    $WD = \{X \mid X \in V \text{ and } BEL(e_1, \ldots, e_n|h) = BEL(e_1'|h) * \ldots * BEL(e_n|h)\}$

- The set of Decomposable Problems D:

    $D = \{X \mid X \text{ is weakly decomposable and } BEL(e_1, \ldots, e_n) = BEL(e_1) * \ldots * BEL(e_n)\}$

- A problem X is holistic if $X \in V - WD$

## Corollary

$D \subset WD \subset V \subset S$

224

## 3. THE RATIONAL INTERPRETATION OF CERTAINTY FACTORS

This section provides a brief account of the definition and interpretation of the Certainty Factors Language, stated originally in [19]. Given a problem $X = (h,e) \in V$, the CF syntax approximates the predictive solution $BEL(h|e)$ through the difference between a measure of increased belief (MB) and a measure of increased disbelief (MD) in the hypothesis h in light of the clue e:

$$-1 \leq CF(h,e) = MB(h,e) - MD(h,e) \leq 1$$

The CF Calculus is a set of operators designed to combine atomic CF's into compound CF's (e.g. compute $BEL(h,a,b)$ from $BEL(h|a)$ and $BEL(h|b)$). In this paper, we will confine the discussion to a subset of this calculus, denoted hereafter (M1)-(M2):

$$MB(h,ab) = \begin{cases} 0 & \text{if } MD(h,ab)=1 \\ MB(h,a) + MB(h,b)*(1-MB(h,a)) & \text{o.w.} \end{cases} \quad (M1)$$

$$MD(h,ab) = \begin{cases} 0 & \text{if } MB(h,ab)=1 \\ MD(h,a) + MD(h,b)*(1-MD(h,a)) & \text{o.w.} \end{cases} \quad (M2)$$

Shortliffe & Buchanan have also suggested a syntactical mapping from Bayesian probabilities to Certainty Factors, defined as follows:

$$MB(h,a) = \begin{cases} 1 & \text{if } P(h)=1 \\ \dfrac{\max(P(h|a),P(h)) - P(h)}{1 - P(h)} & \text{o.w.} \end{cases} \quad (R1)$$

$$MD(h,a) = \begin{cases} 1 & \text{if } P(h)=0 \\ \dfrac{\min(P(h|a),P(h)) - P(h)}{-P(h)} & \text{o.w.} \end{cases} \quad (R2)$$

We term the (R1)-(R2) mapping a _rational interpretation_ for three reasons. First, the mapping is intended to convey a certain degree of descriptive face-validity to the CF language. Second, it relates CF's to a subset of the real interval [0,1] which is consistent with the seven rational postulates of Savage and de Finetti (i.e. the axioms of subjective probability). Third, our notion of a _rational interpretation_ is consistent with Shortliffe & Buchanan who suggest: "_behavior is irrational if actions taken or decisions made contradict the result that would be obtained under a probabilistic analysis of the behavior_" [19, p. 251].

Note that (M1)-(M2) is very different from (R1)-(R2). The former pair of combination rules is the nucleus of the CF calculus, designed to compute the compound strength of belief of two pieces of evidence. The latter pair of definitions is an ex-post Bayesian interpretation of the notion of Certainty Factors that has no operational bearing on CF-based elicitation and updating procedures.

Also note in passing that both (M1) and (M2) appear to convey an (additive) rational sense: if you open the parentheses of (M1) for example, you obtain the sum of $MB(h,a)$ and $MB(h,b)$ minus their multiplicative interaction effect. Nonetheless, the fact that the CF calculus is sensitive to predictive interaction effects does not necessarily extend to the recognition of diagnostic interaction effects, as will be shown below in Lemma 1 (recall that in the context (instantiation) of h being an hypothesis and e being a piece of evidence, $BEL(h|e)$ is termed '_predictive belief_' and $BEL(e|h)$ is termed '_diagnostic belief_').

## 4. THE CF-BAYESIAN ENDOMORPHISM THEOREM

**Lemma 1** If the MB comb. rule (M1) is used to approximate the predictive solution of some problem $X \in V$, then (M1) is mutually consistent with the rational MB interpretation (R1) if and only if X is decomposable.

**Lemma 2** If the MD comb. rule (M2) is used to approximate the predictive solution of some problem $X \in V$, then (M2) is mutually consistent with the rational MD interpretation (R2) if and only if X is decomposable.

**Lemma 3** If the CF calculus (M1)-(M2) is used to approximate the predictive solution of some problem $X \in V$, then (M1)-(M2) is mutually and jointly consistent with the CF rational interpretation (R1)-(R2) if and only if X is weakly decomposable.

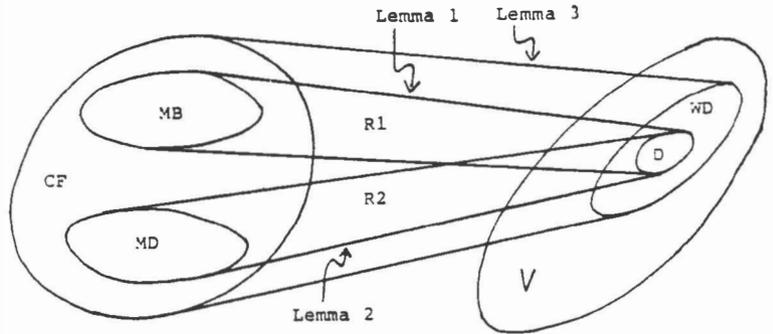

#### THE CF-BAYESIAN ENDOMORPHISM THEOREM

Let CF be the set of all problems that have an approximate predictive solution derived by the CF calculus (M1)-(M2). Let V be the set of all problems that have an approximate predictive solution derived by a Bayesian calculus. Let T be the set of rational interpretation (R1)-(R2). Under these conditions, T is an endomorphic transformation $T : CF \rightarrow WD \subset V$, where WD is the subset of weakly decomposable problems in V.

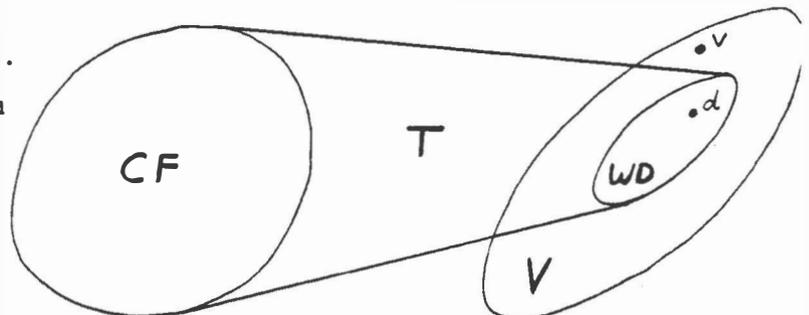

## 5. DISCUSSION

The CF-Bayesian Endomorphism Theorem says that the CF Calculus has a classical rational interpretation if and only if it is restricted to weakly decomposable problems. In the pictorial illustration of the theorem, v is an holistic problem that is outside the rational scope of the CF language (e.g. $v = (h,a,b)$ has a synergistic diagnostic structure, i.e. $BEL(a,b|h) > BEL(a|h)*BEL(b|h)$). At the same time, however, v does have a (complicated) Bayesian solution, by virtue of its membership in V. This dichotomy means that one cannot trade rationality for efficiency, as is sometimes being done in AI. Furthermore, the likelihood that a real-life problem exists in WD is very small, due to the underlying complexity of problems that require expertise [5].

The preceding paragraph implies that most CF-based expert systems (hence, most applied expert systems) are inconsistent with their classical rational interpretation (R1)-(R2). This finding is disturbing, because there exist reported instances where CF-based expert systems have achieved an impressive decision-making performance [8]. There may be (at least) two potential explanations for the disparity between the narrow normative foundation and the de-facto face-validity of the CF language.

First, I argue that experienced Knowledge Engineers intuitively know that the CF-Bayesian Endomorphism Theorem is true, and, in fact, take advantage of it. In particular, designers of complicated expert systems often feel that the more granular the knowledge-base, the higher is the validity of the system [2]. In the context of the present theory, this heuristic amounts to augmenting an evidence / hypotheses inference net with a multitude of sub hypotheses and intermediate states, designed to partition the knowledge-base and achieve a higher degree of granularity. This judicious decomposition is done in an attempt to explicitly account for interaction effects, and, thereby, induce more conditional independence on the evidence / hypotheses space ([5], [13], [16], [22]).

In the context of the present theory, we can describe this practice as follows: when a CF knowledge engineer faces an holistic problem v which is outside the scope of a rational interpretation, he or she first modifies the diagnostic structure of the original problem, thus creating a transformation from v to d∈WD, which is a rational CF territory. If there exists a problem d whose diagnostic structure is indeed a plausible (weak) decomposition of v, a CF-based system applied to d is likely to provide a (close) rational belief representation to v as well.

The second explanation of the CF descriptive / normative contrast may be that the original rational interpretation of Certainty Factors (R1)-(R2) is subject to doubt, as was also noted by Heckerman [12]. In other words, it seems that the notion of CF's is indeed a novel idea that deserves a serious look, especially on practical and descriptive grounds. Indeed, the fact that the CF movement has been going strong for more than a decade in spite of its unrealistically narrow rational interpretation suggests that the model is basically powerful although its normative foundation is weak. Hence, future basic research is needed to explore new interpretations to the CF language that will be more plausible on rational, cognitive, and philosophical grounds.

**Conjecture C1 Revisited** We now turn to the casual conjecture C1, which attributes the impracticality of the Bayesian approach vis a vis the CF language to the fact that real life domains of expertise are not statistically independent. This statement, in my view, is based on a semantic rather than a substantive argument. In particular, I have suggested elsewhere that statistical independence is not directly expressible in the CF formalism [16]. This is consistent with Shafer & Tversky, who observe that some mathematical properties are not translatable from one probabilistic language to another [18]. However, the fact that a particular characteristic of the world cannot be described in a certain language does not necessarily imply that this characteristic in nonexistent.

The statistical independence phenomenon is an attribute of nature which stands aloof from the CF/Bayesian debate. To clarify this distinction, we may use an analogy from physics. The presence or absence of statistical independence is a unique property of a domain of expertise just as the mass is a unique physical property of a brick. Notwithstanding the mass uniqueness, the weight of the brick varies with different scales (or on different planets). Thus an absolute unique property of nature may be mapped onto different manifestations under different circumstances. Similarly, the manifestation of the independence property may be explicit in some probabilistic languages and vague or even null in others. The crispness of this expression should be construed as a property of the language, not a property of nature.

**Conjecture C2 Revisited** The C2 conjecture suggests that both the CF and the Bayesian languages are special cases of the Dempster-Shafer Theory of Evidence. Although this premise is indeed correct, this truth is quite different from its popular interpretation. That the Bayesian design is a special case of the Dempster-Shafer model is a trivial corollary that can be found in [19]. Similarly, Gordon & Shortliffe gave a Shaferian belief interpretation to the CF calculus. They then proceeded to conclude that "the Dempster-Shafer combination rule includes the Bayesian and the CF functions as special cases [11, p. 273]."

In my view, the popular interpretation of this correct argument is as follows:

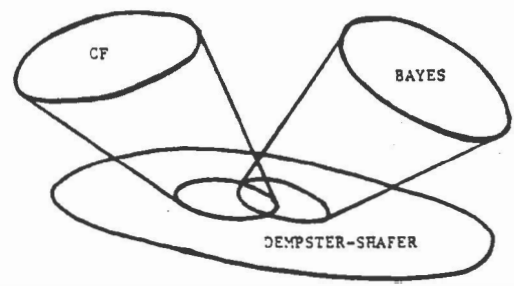

226

However, in light of the CF-Bayesian Endomorphism Theorem, a more accurate description of the CF/Bayesian/Shafer relationships is as follows:

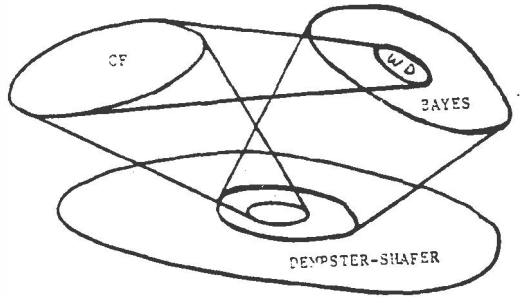

This view suggests a new Bayesian interpretation of the controversial subadditivity property of CF's that Gordon & Shortliffe alluded to in [10]. This pragmatic interpretation, which is beyond the scope of this paper, may be found in [16]. In short, our analysis indicates that within the subset WD, CF's are remarkably Bayesian after all. Outside the subset WD, the CF language may have a variety of free-form and appealing descriptive interpretations. At the same time, those ad-hoc interpretations will not be accountable or testable on rational grounds, if only for the reason that a rational CF interpretation currently does not exist outside WD. Of course, this restriction may be lifted if the original interpretation (R1)-(R2) is modified or extended in order to cover a larger superset of WD.

## 6. IMPLICATIONS ON KNOWLEDGE ENGINEERING

Several authors have stressed the fact that most real-life domains of expertise include problems that are *'dependant'* (Pearl), *'adjacent'* (Davis), or *'holistic'* (Schocken). Clearly, all those definitions basically imply that the diagnostic structure of realistically complex problems is not weakly decomposable. This observation has far-reaching computational implications which may be summarized in the following proposition, which has not yet been proven rigorously:

Proposition: If a problem $(h, e_1 \ldots e_n)$ is holistic, then the computation of its predictive solution $BEL(h|e_1, \ldots, e_n)$ is $NP$-Complete.

Heuristic argument: This proposition is based on the intuition that an n-cities Travelling Salesman Problem (TSP) is reducible to the Integer Programming (IP) formulation of $BEL(h|e_1, \ldots, e_n)$, provided that such plausible formulation exists. The objective function of the IP model will be based on a logarithmic transformation of $BEL(h|e_1, \ldots, e_n)$ which amounts to a linear combination of all the possible interaction effects within subsets of the $(h, e_1, \ldots, e_n)$ space. Each individual interaction effect is multiplied by an integer 0-1 variable that determines whether or not the interaction obtains. Furthermore, it is felt that the typical TSP constraint designed to avoid a sub-tour among k<n cities can be mapped onto the IP constraint that given that X is holistic, the computation of BEL cannot afford to disregard a dependency of degree k within $(h, e_1, \ldots, e_k)$. Those $2^n$ logical constraints will force the appropriate 0-1 variables to be set to 0 or

1, thus determining whether or not the respective interaction effect should enter the computation of $BEL(h|e_1, \ldots, e_n)$ in the objective function.

We now return to the implications of the CF-Bayesian Endomorphism Theorem on knowledge engineering in light of the proposition just presented. In particular, we wish to focus on the key question that we ought to address, namely: how can a rule-based expert system compute the predictive solution of an holistic problem. Basically, there seem to be three alternative options:

1. Apply a rule-based algorithm as though the problem in not holistic:

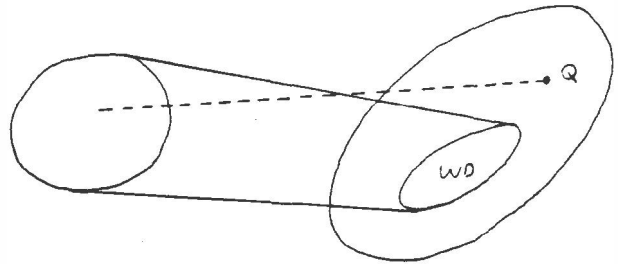

2. Devise a new, *non* rule-based holistic algorithm:

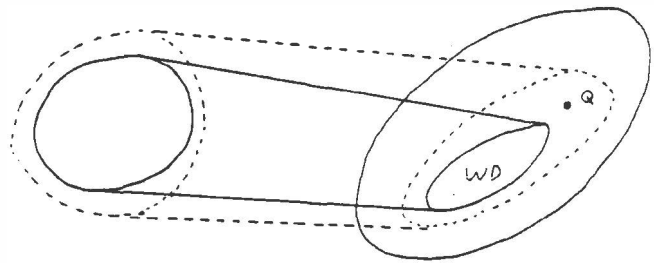

3. Transform the holistic problem Q into a more complicated problem Q' that nonetheless is (roughly) weakly decomposable. Then apply a rule-based algorithm to Q'.

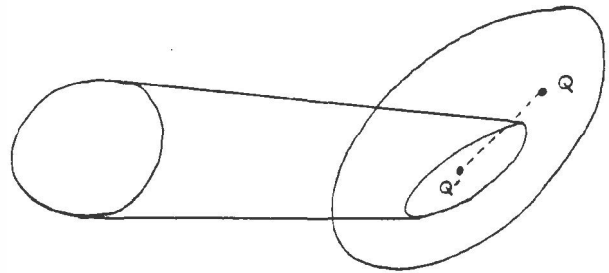

Option 1, which basically amounts to fudging, is, in my opinion, the leading practice among practitioners of rule-based expert system shells (this intuitive proposition is not supported by any firm empirical evidence). Moreover, the CF-Bayesian Endomorphism Theorem shows that it doesn't really matter if you use a decomposable Bayesian or a CF approach; both fail to handle holistic diagnostic structures, although the latter implicitly claims that it does.

Option 2 is too good to be true. The existence of such algorithm would imply that a polynomial algorithm was found to the TSP. In light of Cook's Theorem [6], this will imply that $P \ne NP$, amounting to the most staggering finding in Complexity Theory, and a very unlikely one.

This leaves us with the pragmatic option 3, in which, in my opinion, we ought to invest our intellectual efforts. If we manage to construct a plausible decomposition Q' of Q, then we can safely apply a rule-based predictive solution to Q', whose distance from the optimal solution is $|BEL(Q'|E')-BEL(Q|E)|$. Some ideas about how such decomposition can be carried out may be found in the seminal work of Pearl [13] as well as in Charniak's notion of 'intermediate states' [5].

If we view the optimal solution of Q as a complete combinatorial Bayesian design, and the rule-based solution of Q' as a heuristic solution of Q, we may bring upon some very strong findings from the probabilistic analysis of the TSP that followed the 1971 $NP$-completeness revolution. For example, there is a greedy algorithm that solves the TSP in polynomial time, giving a solution that may not be optimum but is guaranteed to be no worse than twice the optimum path (due to Christofides of the Imperial College of Science and Technology in London). Furthermore, if some very plausible assumptions are made regarding the layout of the cities (viz, the topology of the diagnostic structure of the problem), this error can become as small as 5% (due to Beardwood of Oxford, and also Karp).

## 7. CONCLUSION

This paper presents some analytical results that are emerging from an elaborate research program on uncertainty in AI. One important aspect of this undertaking that was not covered here is the investigation of the semantical (cognitive) dimension of the CF and the Bayesian languages. This effort, which involves a large-scale experiment that utilizes experts as well as (CF and Bayesian) expert systems as subjects, is already underway. We believe that this research will improve significantly our understanding of the philosophical foundations of expert systems as well as the practicality of various approaches to uncertainty in AI. This effort may promote the formulation of a synthetic approach to Knowledge Engineering, i.e. one that combines the attractive descriptive features of the CF language with the normative rigor of a Bayesian design. It is hoped that this approach will strike a balance between preserving the intuitive element of human expertise, and, at the same time, enforcing a certain degree of normative rationality.

## REFERENCES

1. Adams J.B. 'Probabilistic Reasoning and CF's' in 'Rule-Based Expert Systems.' B.G. Buchanan & E.H. Shortliffe (Eds.), Addison-Wesley, 1984.

2. Barr A. & Feigenbaum E.A. (Eds.). 'The Handbook of Artificial Intelligence.' Los Altos, CA: William Kaufmann, 1981, p 195.

3. Buchanan B.G. & Shortliffe E.H. 'Uncertainty and Evidential Support.' in 'Rule-Based Expert Systems.' Buchanan B.G. & Shortliffe E.H. (Eds.), Addison-Wesley, 1984, p. 215.

4. Carnap R. 'Logical Foundations of Probability.' Chicago: University of Chicago Press.

5. Charniak E. 'The Bayesian Basis of Common Sense Medical Diagnosis.' Proc. of the National Conference in Artificial Intelligence, 1983, 3, pp. 70-73.

6. Cook S.A. 'The Complexity of Theorem Proving Procedures.' Proc. 3rd ACM Symp. On The Theory of Computing, ACM (1971).

7. Duda R.O., Gaschnig J., & Hart P. 'Model Design in the Prospector Consultant System for Mineral Exploration.' In 'Expert Systems in the Microelectronic Age.' Donald Michie (Ed.).

8. Duda R.O & Shortliffe E.H. 'Expert Systems Research.' Science 220: pp. 261-268

9. Fischhoff. 'Debiasing.' in 'Judgment under Uncertainty, Heuristics and Biases.' D. Kahneman, P. Slovic, & A. Tversky (Eds.), Cambridge, Cambridge University Press, 1982.

10. Gordon J. & Shortliffe E.H. 'The Dempster Shafer Theory of Evidence.' in 'Rule-Based Expert Systems.' Buchanan B.G. & Shortliffe E.H. (Eds.), Addison-Wesley, 1984.

11. Kahneman D. & Tversky A. 'Intuitive Prediction: Biases and Corrective Procedures.' Management Science, 1979, 12, 313-327.

12. Heckerman D. 'Probabilistic Interpretations for Mycin's Certainty Factors.' Medical Computer Science Group, TC135, Stanford University Medical Center.

13. Pearl J. 'Fusion, Propagation, and Structuring in Bayesian Networks.' Technical Report CSD-850022, U.C.L.A., April, 1985.

14. Prade H. 'A Synthetic View of Approximate Reasoning Techniques.' IJCAI 8, Karlsruhe, W. Germany, 1983.

15. Savage L.J. 'The Foundations of Statistics.' Wiley, 1954.

16. Schocken S. 'On the Underlying Rationality of Expert Systems: the Analytical Ground.' DS Working Paper, Decision Sciences Department, The Wharton School of the University of Pennsylvania, 04-06-86.

17. Shafer G. 'A Mathematical Theory of Evidence.' Princeton University Press, 1976.

18. Shafer G. & Tversky A. 'Languages and Designs for Probability Judgment.' Cognitive Science 9, 1985.

19. Shortliffe E.H. & Buchanan B.G. 'A Model of Inexact Reasoning in Medicine.' in 'Rule-Based Expert Systems.' Buchanan B.G. & Shortliffe E.H. (Eds.), Addison-Wesley, 1984, p. 239.

20. Tversky A. & Kahneman D. 'Judgment Under Uncertainty - Heuristics and Biases.' Science, Vol. 185, September 1974.

21. Wallsten T.A. & Budescu D.V. 'Encoding Subjective Probabilities: a Psychological and Psychometric Review.' Management Science, Vol. 29, No. 2, Feb. 1983.

22. Winter C.L. & Girse R.D. 'Evolution & Modification of Probability in Knowledge Bases.' The 2nd Conference on AI Applications, 1985.